\documentclass[final]{cvpr}

\usepackage{times}
\usepackage{epsfig}
\usepackage{graphicx}
\usepackage{amsmath}
\usepackage{amssymb}

\usepackage[pagebackref=true,breaklinks=true,colorlinks,bookmarks=false]{hyperref}

\def\cvprPaperID{5636} 
\def\confYear{CVPR 2021}

\begin{document}

\title{Direct Depth Learning Network for Stereo Matching}

\author{Hong Zhang\textsuperscript{1}, Haojie Li\textsuperscript{2}, Shenglun Chen\textsuperscript{3}, Tiantian Yan\textsuperscript{4}, Zhihui Wang\textsuperscript{5}\\
Dalian University of Technology\\
{\tt\small \{jingshui\textsuperscript{1},1936902534\textsuperscript{3}\}@mail.dlut.edu.com} \\
{\tt\small\{hjli\textsuperscript{2},zhwang\textsuperscript{5}\}@dlut.edu.com, 2352888210\textsuperscript{4}@qq.com}
\and
Guo Lu\\
Beijing Institute of Technology\\
{\tt\small sdluguo@gmail.com}
\and
Wanli Ouyang\\
The University of Sydney\\
{\tt\small wanli.ouyang@sydney.edu.au}
}

\maketitle

\begin{abstract}
  Being a crucial task of autonomous driving, Stereo matching has made great progress in recent years. Existing stereo matching methods estimate disparity instead of depth. They treat the disparity errors as the evaluation metric of the depth estimation errors, since the depth can be calculated from the disparity according to the triangulation principle. However, we find that the error of the depth depends not only on the error of the disparity but also on the depth range of the points. Therefore, even if the disparity error is low, the depth error is still large, especially for the distant points. In this paper, a novel Direct Depth Learning Network (DDL-Net) is designed for stereo matching. DDL-Net consists of two stages: the Coarse Depth Estimation stage and the Adaptive-Grained Depth Refinement stage, which are all supervised by depth instead of disparity. Specifically, Coarse Depth Estimation stage uniformly samples the matching candidates according to depth range to construct cost volume and output coarse depth. Adaptive-Grained Depth Refinement stage performs further matching near the coarse depth to correct the imprecise matching and wrong matching.  To make the Adaptive-Grained Depth Refinement stage robust to the coarse depth and adaptive to the depth range of the points, the Granularity Uncertainty is introduced to Adaptive-Grained Depth Refinement stage. Granularity Uncertainty  adjusts the matching range and selects the candidates' features according to coarse prediction confidence and depth range. We verify the performance of DDL-Net on SceneFlow dataset and DrivingStereo dataset by different depth metrics. Results show that DDL-Net  achieves an average improvement of $25\%$ on the SceneFlow dataset and $12\%$ on the DrivingStereo dataset comparing the classical methods. More importantly, we achieve state-of-the-art accuracy at a large distance.
\end{abstract}
\section{Introduction}
Depth estimation is pivotal to a variety of high-level tasks in computer vision, such as autonomous driving, robot navigation \cite{Semi-directSLAM}, object detection and recognition \cite{DBLP:conf/cvpr/WangCGHCW19,DBLP:conf/iccv/MaWLZOF19}. Stereo Matching (SM) is one of the most important passive depth estimation methods, which estimates depth by utilizing the \emph{Triangulation principle}, $Depth = \frac{B\times f}{disparity}$. $B$ and $f$ are the baseline and focal length of the calibrated stereo camera system, respectively. 

It is generally supposed that more accurate disparity means more accurate depth. Therefore, existing SM methods \cite{SGM,DBLP:journals/ijcv/ScharsteinS02,DBLP:journals/pami/BrownBH03} focus on improving the performance of disparity estimation. They perform matching on rectified stereo images and obtain accurate disparity firstly. Then the depth is calculated according to the triangulation principle. Recent deep learning based stereo matching methods \cite{Stereonet,DBLP:conf/cvpr/LuoSU16,GCNet} map the rectified binocular images to feature space through shared CNN, and construct disparity-based cost volume in feature space. Then the cost volume is optimized by cost aggregation network, such as 3D convolution network \cite{PSMNet,GwcNet}, multi-scale fusion\cite{AANet}, and learnable semi-global propagation \cite{GANet,DBLP:conf/cvpr/SekiP17}. These learning methods bring in higher disparity accuracy on the existing evaluation criteria. 

However, we argue that the performance metrics for stereo matching, such as the end-point error (EPE), are insufficient to evaluate the accuracy of depth in complex driving scenes \cite{DBLP:conf/cvpr/YangSHDSZ19}. As shown in Equation \ref{error}, the accuracy of depth not only depends on the disparity errors, but also relies on the ground-truth depth. Therefore, even if the disparity error is low, the depth error is still large when $depth_{gt}$ is large.
\begin{equation}
\begin{split}
  depth_{error} = |\frac{B\cdot f}{dis_{gt}}-\frac{B\cdot f}{dis_{pred}}| \\
= depth_{gt}\cdot \frac{dis_{error}}{ dis_{pred}}, 
\end{split}
\label{error}
\end{equation}
where $dis_{gt}$, $dis_{pred}$ and $dis_{error}$ are ground-truth disparity, predicted disparity and disparity error, respectively. $depth_{gt}$ is the ground-truth depth. 

Consequently, a key issue is to design an effective framework that directly output more accurate depth rather than disparity. The predicted depth is required to adapt to tasks such as autonomous driving and object detection in a complex environment, which need accurate depth at both near and far. In this work, a Direct Depth Learning Network (DDL-Net) is proposed to improve the accuracy of depth estimation rather than disparity estimation. 
DDL-Net contains two stages: the Coarse Depth Estimation stage (CDE) and the Adaptive-Grained Depth Refinement stage (AGDR). 

CDE estimates a coarse depth map to narrow the matching range of AGDR stage. We construct the cost volume in CDE by uniformly selecting matching candidates according to depth instead of disparity adopted by most of the existing SM methods \cite{GwcNet,PSMNet}. This is because sampling the matching candidates uniformly according to disparity could lead to the problem that the errors of depth estimation grow quadratically with depth. Therefore, the accuracy of the points at large distance is improved significantly by depth-based cost volume.

AGDR performs further matching in the narrow matching range like multi-stage matching methods. However, the existing fine matching methods can not be directly employed in depth estimation. The fine matching range of methods like  \cite{DBLP:conf/cvpr/TonioniTPMS19,DBLP:conf/cvpr/YinDY19} depends on the coarse prediction without considering large errors. Besides, the determination of matching granularity does not consider the fact that farther points need denser matching. Here we unify the fine matching range (it should be wider for large errors), and granularity(it should be denser for farther depth) as matching granularity. According to the above analysis, the matching granularity should be adaptive to the depth and the confidence of the coarse prediction. We propose a Granularity Uncertainty (GU) to adjust the matching granularity of the fine matching. GU contains two parts: (1) Scale Uncertainty to adjust the matching range by changing the scale of the offset (the offset is the distance from the coarse prediction to ground-truth); (2) Feature Uncertainty for the matching features which adaptively selects the candidates points for matching. The Feature Uncertainty varies with the Scale Uncertainty. The ablation study demonstrates that the accuracy at both far and near is improved by introducing GU.

In sum, the contributions of the paper are two-fold:
\begin{itemize} 
\item A Direct Depth Learning Network(DDL-Net) is designed to directly improve the accuracy of depth rather than disparity in this work. The depth accuracy at larger distance is significantly improved.
\item The Granularity Uncertainty guided adaptive-grained depth refinement is proposed to make the matching granularity adapt to the depth and the confidence of the coarse prediction. GU makes the DDL-Net not only perform best at a large distance but also maintain the accuracy at a small distance.
\end{itemize} 

We conduct experiments on SceneFlow \cite{Sceneflow} dataset and DrivingStereo \cite{drivingstereo} dataset, and the experimental results show that DDL-Net  achieves an average improvement of $25\%$ on the SceneFlow dataset and $12\%$ on the DrivingStereo dataset than the classical methods. More importantly, we achieve state-of-the-art accuracy at a large distance.  

\begin{figure*}[h]
\begin{center}
\includegraphics[width=0.8\textwidth]{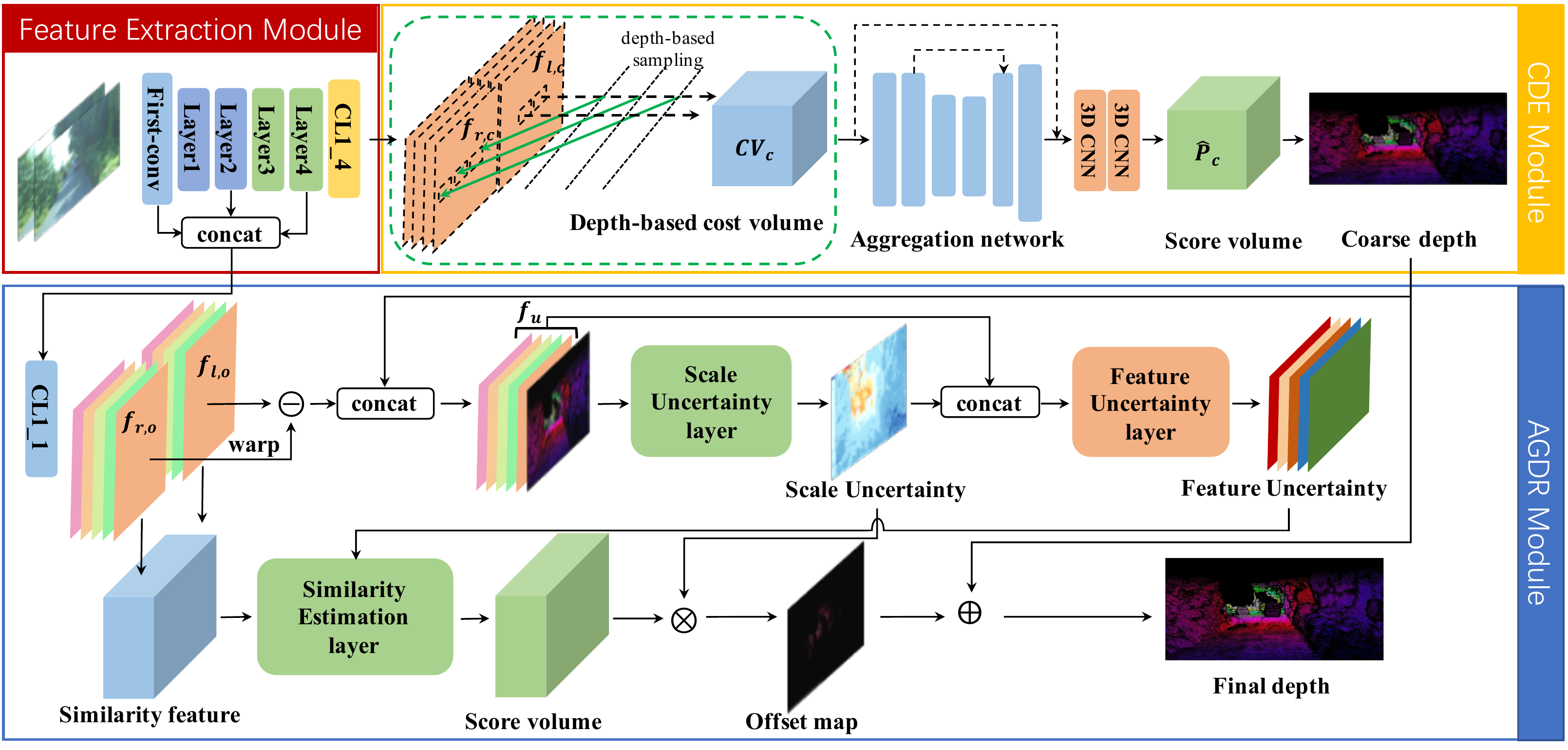}
\end{center}
   \caption{The framework of the proposed DDL-Net. $\ominus$ : element-wise subtraction; $\oplus$ : element-wise addition; $\otimes$ : the operation of Equation \ref{offset}. Firstly, in the Feature Extraction module, image pairs are mapped into features  $f_{l,c},f_{r,c}$ and $f_{l,o},f_{r,o}$. Then the low resolution features $f_{l,c},f_{r,c}$ are sent to the CDE module to construct depth-based cost volume and output coarse depth. Finally, $f_{l,o},f_{r,o}$ and the coarse depth are fed to the AGDR module to perform GU (SU and FU) guided further matching and output the accurate depth.}
\label{fig:framework}
\end{figure*}

\section{Related work}
\paragraph{Disparity estimation} This section reviews recent end-to-end supervised deep learning stereo matching methods.

2D CNN based methods, such as DispNetC \cite{Sceneflow}, CRL \cite{DBLP:conf/iccvw/PangSRYY17} and iRes-Net \cite{DBLP:conf/cvpr/LiangFGLCQZZ18} are end-to-end trainable disparity estimation network. They form a low-resolution 3D cost volume by calculating the cosine distance of each unary feature with their corresponding unary from the opposite stereo image across each disparity level. Then the 3D cost volume is put into 2D CNN with supplementary features for disparity regression. After that,  SegStereo \cite{DBLP:conf/eccv/YangZSDJ18} and EdgeStereo \cite{DBLP:journals/corr/abs-1903-01700} design multiple tasks frameworks for the disparity regression task. The former introduces semantic information in the refinement stage and the latter applies edge information for guiding disparity optimization. 

 Cost aggregation network-based methods study how to optimize the low-resolution 4D volume to obtain more accurate similarity scores in the low-resolution 3D cost volume and output a better disparity map accordingly. Yu \emph{et al.} \cite{DBLP:conf/aaai/YuWWJ18} propose an explicit cost aggregation sub-network to provide better contextual information. PSM-Net \cite{PSMNet} introduces a pyramid pooling module for incorporating global context information into image features, and stacked 3D CNN hourglasses to extend the regional support of context information in cost volume. In order to make full use of the features, Gwc-Net \cite{GwcNet} builds the cost volume by concatenating the cost volume constructed in different ways. GA-Net \cite{GANet} proposes two new neural net layers to capture the local and the whole-image cost dependencies and to replace the 3D convolutional layer. AA-Net \cite{AANet} proposes a sparse points based intra-scale cost aggregation method to achieve fast inference speed while maintaining comparable accuracy. 

Multistage matching methods like \cite{DBLP:conf/cvpr/TonioniTPMS19,DBLP:conf/cvpr/YinDY19} first obtain a coarse disparity and then perform residual disparity search from the neighbor of the current disparity by constructing a partial cost volume. DeepPruner \cite{DeepPruner}  develops a differentiable PatchMatch module to provide a learned narrow search range that allows discarding most disparities without requiring full cost volume evaluation in the second stage. 

These stereo matching methods directly estimate disparity and achieve remarkable accuracy on the disparity. However, the purpose of stereo matching is to estimate accurate depth other than disparity. Although they perform excellently on disparity estimation, they all have poor performance on depth accuracy,  especially at large depth region. In this work, we focus on depth accuracy and propose a novel two-stage framework. There are three main differences compared with the existing stereo matching methods: (1) It directly outputs depth and is supervised by depth; (2) It constructs cost volume by sampling matching candidates on depth instead of disparity; (3) a novel GU is designed in the fine matching stage instead of matching near the coarse prediction between a fixed range.

\paragraph{Depth estimation} Although existing stereo matching methods do not focus on the accuracy of depth estimation, there are many depth estimation methods in other fields. 

Monocular depth estimation \cite{DBLP:conf/3dim/LainaRBTN16,DBLP:conf/cvpr/FuGWBT18,DBLP:conf/wacv/HuOZO19,MonocularCRF} methods estimate depth from single image, which rely primarily on prior knowledge rather than geometric knowledge. Although there are some self-supervised methods utilize the other image \cite{DBLP:conf/cvpr/GodardAB17} of binocular images or adjacent frames of monocular sequence \cite{DBLP:conf/cvpr/YangSWC20} for supervision, the estimation process still relys on one image. Comparatively speaking, it is more reliable by utilizing geometric information in binocular images to estimate the depth.

Recently, stereo 3D object detection tasks \cite{DBLP:conf/cvpr/SunCXZJZB20}\cite{DBLP:conf/cvpr/Chen0SJ20} also introduce binocular depth estimation instead of directly using point cloud. Among them, DSGN \cite{DBLP:conf/cvpr/Chen0SJ20} constructs a plane-sweep volume for 3D object detection by concatenating the left image feature and the reprojected right image feature at equally spaced depth interval. However, they focus on object detection and stays in the coarse phase for depth estimation.

Multi-view depth estimation (MVS) \cite{DBLP:conf/cvpr/YangMAL20}\cite{DBLP:conf/cvpr/GuFZDTT20}\cite{DBLP:conf/cvpr/YangMAL20} aims to reconstruct the 3D model from a set of images captured by a camera. MVS constructs cost volume on depth plane sampled according to the parameter between multiple views. In these methods, UCS-Net \cite{UCSNet} uses the uncertainty of the previous prediction to adjust the matching range of the refinement stage, which is similar to the work in this paper. However, we are different from UCS-Net that we focus on SM and the application scene, such as autonomous driving, is more complex than UCS-Net. First, the depth range is larger in autonomous driving while small in 3D module reconstruction. Second, there are a diversity of objects and many interferences (such as the sky) in the road scene of SM while MVS only needs to estimates the depth of a single module. Therefore, We need to adjust the matching granularity according to depth value, and the UCS-Net does not need. The AGDR in this paper is designed to deal with this problem. 

\section{Method}

Given a pair of rectified stereo images, $I_1$ and $I_2$, we aim to estimate a dense depth map using stereo geometry information. An overview of our approach is given in Figure \ref{fig:framework}. Firstly, based on the input stereo images $I_1$ and $I_2$, we utilize the feature extraction module to produce the low resolution feature maps $f_{l,c},f_{r,c}$ and the original feature maps  $f_{l,o},f_{r,o}$. Then in the proposed CDE module, we use features $f_{l,c}$ and $f_{r,c}$ to construct depth-based cost volume and estimate the coarse depth. Finally, we further refine the coarse depth map in the AGDR module under the guidance of SU and FU, obtaining a more accurate depth map.

\subsection{Feature extraction}
Feature extraction module provides $\frac{1}{4}$ resolution feature maps $f_{l,c},f_{r,c}$ for coarse depth estimation, and  original resolution feature maps $f_{l,o},f_{r,o}$ with multiple receptive fields for adaptive depth refinement, respectively.
As shown in Figure \ref{fig:framework}, We adopt a ResNet-like network utilized in PSM-Net \cite{PSMNet} for feature extraction. The feature maps output from \emph{Layer4} is compressed by \emph{CL1\_4} to form 32-channel $f_{l,c}$ and $f_{r,c}$. For $f_{l,o}$ and $f_{r,o}$, we first concatenate the upsampled features output from \emph{Layer4}, \emph{Layer2} and the full resolution features output from \emph{First-conv}, and then compress the concatenated features with \emph{CL1\_1} to 32 channels.

\subsection{Coarse depth estimation}
In order to reduce computational cost and satisfy the matching density at different distances, we perform coarse depth estimation to narrow the matching range. 

Given $f_{l,c},f_{r,c}$, the next step is to construct cost volume for coarse depth inference in the left image.
In this paper, in order to solve the imbalance of the depth accuracy of the near and far, we construct the cost volume by matching the candidates uniformly sampled according to depth range.
Specifically, we first uniformly sample D fronto-parallel planes across the entire depth range. Suppose the matching range is $[d_{min},d_{max}]$, then $d_i=d_{min}+i\cdot(d_{max}-d_{min})/D$, where $i \in \{0,1,2,\cdots , D-1\}$ represents the $(i+1)$-th sampled plane whose normal is the principal axis of the stereo camera system. Given the intrinsic parameter focal length $f$ and extrinsic parameter baseline $B$ of binocular camera system, the $(i+1)$-th matching candidates $C_i(x,y)$  of the reference point $(x,y)$ can be obtained by Equation \ref{candidates}. 
\begin{equation}
   C_i(x,y) = \frac{f\cdot B}{d_i}, i \in {0,1,2,\cdots , D-1}.
\label{candidates}
\end{equation}

Secondly, a 4D cost volume $CV_{c}$ based on depth is constructed:
\begin{equation}
 CV_{c}(x,y,i ) = <f_{l,c}(x,y),f_{r,c}( C_i(x,y))>,
 \label{costvolume}
\end{equation}
where $<\cdot>$ represents concatenation operation. As shown in Figure \ref{fig:framework}, $CV_{c}$ will then be sent to an aggregation network, obtaining the matching score volume $\hat{P}_{c}$.

Thirdly, $\hat{P}_{c}$ is upsampled by trilinear interpolation and normalized by $Softmax$ operation obtaining $P_{c}$. The depth estimation for each pixel $(x,y)$ is computed as
\begin{equation}
    depth_{coarse}(x,y) = \sum_{i=0}^{D-1}(i\cdot P_{c}(x,y,i)),
    \label{cosrseD}
\end{equation}
where $depth_{coarse}(\cdot)$ represents the corresponding coarse depth map. 

\subsection{Adaptive-grained depth refinement}
The proposed AGDR obtains accurate depth map according to Equation \ref{finaldepth} by estimating the offset between the coarse depth and the accurate depth,
\begin{equation}
depth = depth_{coarse}+offset,
\label{finaldepth}
\end{equation}
where $offset$ is calculated from the GU (SU and FU) guided fine matching process. 

As shown in Figure \ref{fig:framework}, firstly, we subtract $f_{l,o}(x,y)$ from the warped $f_{r,o}(x,y)$ (it is warped according to the coarse depth) to obtain the uncertainty feature $f_u$. Secondly, the uncertainty feature and the coarse depth are concatenated and fed to the Scale Uncertainty layer to estimate SU, which is employed to adjust the matching range. SU and the uncertainty feature are concatenated and sent to the Feature Uncertainty layer to calculate FU. FU is used to select the appropriate features when calculating the similarity scores. Thirdly, we construct the similarity features with $f_{l,o}$ and $f_{r,o}$, and send it to the Similarity Estimation layer to calculate the score volume under the guidance of FU. Finally, the offset is obtained by Equation \ref{offset},
\begin{equation}
\begin{split}
   &offset = \\
   & \sum_{i=-2}^{2}(SU(x,y)\times i\times s(f_{l,o}(x,y),f_{r,o}(x^{\prime}+i,y)|FU)),  
\end{split}
\label{offset}
\end{equation}
where $s(f_{l,o}(x,y),f_{r,o}(x^{\prime}+i,y)|FU)$ is a similarity score of the score volume, which represents the similarity between the FU selected features $f_{l,o}(x,y)$ and $f_{r,o}(x^{\prime}+i,y)$.


\paragraph{Feature uncertainty (FU)}FU measures the importance of the features provided for fine matching, which varies with SU. We use FU to select the suitable candidates by multiplying it as a weight over the provided features.

We first encode FU in the feature representation and take it into account during matching. Suppose $z_{l}(x,y)$ and $z_{r}(x,y)$ are the latent features of the input pixels $I_{l}(x,y)$ and $I_{r}(x,y)$, respectively. Inspired by \cite{faceEmbedding}, the distribution of $z_{l}(x,y)$ and $z_{r}(x,y)$ can be modeled as a Multivariate Gaussian distribution
\begin{equation}
\begin{split}
    &p(z(x,y)|I(x,y)) = \\ &\mathcal{N}(z(x,y); f_{o}(x,y),\sigma^{2}(x,y)E),
\end{split}
\end{equation}
where $(x,y)$ represents the pixels' position. $ f_{o}(x,y)$ and $\sigma(x,y)$ are C-dimensional vector. $E$ is the identity matrix. Here we only consider a diagonal covariance matrix to reduce the complexity of feature representation. 

To judge whether $I(x,y)$ and $I(x^{\prime},y)$ represent the same point in the world coordinate system, the probability of $z_l(x,y)$ = $z_r(x^{\prime},y)$ is requried to be calculated. Given $\Delta z=|z_l(x,y)-z_r(x^{\prime},y)|=0$, then the probability $p(z_l(x,y)=z_r(x^{\prime},y))$ is equivalent to the density value $p(\Delta z=0)$. The $c$-th dimension of $\Delta z$, \emph{i.e.} $\Delta z^c$ is the subtraction of two Gaussian variables,which means: 
\begin{equation}
\begin{split}
    \Delta z^{c} &\sim \mathcal{N}(f_{l,o}^{c}(x,y)-f_{r,o}^{c}(x^{\prime},y), \\
    &\sigma_{l,(x,y)}^{2(c)}+\sigma_{r,(x^{\prime},y)}^{2(c)}-2\Sigma),
\end{split}
\end{equation}
where $\Sigma$ is the covariance matrix of  $z_l(x,y)$ and $z_r(x^{\prime},y)$.

Then, $p(\Delta z =0)$ is represented as
\begin{equation}
    p(\Delta z =0)=\sum_{c=0}^{C}p(\Delta z^c=0).
\end{equation}

Taking the logarithm of the probability distribution to represent the similarity distribution, the similarity score between the two features can be represented as 
\begin{equation}
    \begin{split}
    &s(f_{l,o}(x,y),f_{r,o}(x^{\prime},y)) \\
    &= log(p(\Delta z=0)) \\
    &= -\frac{1}{2}\sum_{c=0}^{C}(\frac{(f_{l,o}^{c}(x,y)-f_{r,o}^{c}(x^{\prime},y))^2}{\sigma_{l,(x,y)}^{2(c)}+\sigma_{r,(x^{\prime},y)}^{2(c)}-2\Sigma}\\&
    +log(\sigma_{l,(x,y)}^{2(c)}+\sigma_{r,(x^{\prime},y)}^{2(c)}-2\Sigma))-\frac{C}{2}log2\pi.
    \end{split}
    \label{similarity}
\end{equation}

We analyze Equation \ref{similarity} by dividing it into three terms: 

    (1) $-(f_{l,o}^{c}(x,y)-f_{r,o}^{c}(x^{\prime},y))^2$ : the similarity between the two features $f_{l,o}^{c}(x,y)$ and $f_{r,o}^{c}(x^{\prime},y)$, and larger means more similar.
    
    (2) $\sigma_{l,(x,y)}^{2(c)}+\sigma_{r,(x^{\prime},y)}^{2(c)}-2\Sigma$ : the feature uncertainty, which varies with the value and confidence of the coarse depth. For simplification, the uncertainty is represented as $\sigma_{f,(x,y,x^{\prime})}^{2(c)}$. To avoid zero denominator, we define $\frac{1}{\sigma_f^2}$ as FU. 
   
    (3) $-log(\sigma_{l,(x,y)}^{2(c)}+\sigma_{r,(x^{\prime},y)}^{2(c)}-2\Sigma)-\frac{C}{2}log2\pi$ : a penalty term utilized to punish an uncertain feature. We do not consider this item in this work.

Next, we introduce FU to the network according to the Equation  \ref{similarity}.
 
 (1) Similarity features: First, assuming $(x^{\prime},y)$ is the corresponding position on $I_r$ of the point $I_l(x,y)$ under the coarse depth, we select the candidate features at position $\{x^{\prime}-2,x^{\prime}-1,x^{\prime},x^{\prime}+1,x^{\prime}+2\}$ of $f_{r,o}$ for fine matching. It is worth noting that these candidate features do not represent features of pixels $\{x^{\prime}-2,x^{\prime}-1,x^{\prime},x^{\prime}+1,x^{\prime}+2\}$ on $I_r$, because the features $f_{l,o},f_{r,o}$ are multiple receptive fields features which contain not only the features of current position, but also the features of multiple pixels in the receptive fields. Then, we 
send the similarity features $(f_{l,o}-f_{r,o})$ to two 3D convolutions of the Similarity estimation layer to aggregate context information, obtaining $f_{s}$(a 4D similarity feature volume of $C\times5\times H\times W$). Finally, we unfold the 4D volume along similarity dimensions, obtaining  $3C\times5\times H\times W$ volume $f^{\prime}_{s}$, to provide larger receptive field for each matching position.


(2) Feature uncertainty: We concatenate the $f_u$ and \emph{scale map} (which are introduced in the \emph{Scale uncertainty} part), then send it to three 2D convolutions (Feature Uncertainty layer) to estimate FU, the shape of which is $3C\times5\times H\times W$.

(3) Similarity volume: $f_{s}^{\prime}$ is selected by multiplying FU. We send the similarity features selected by FU to another two 3D convolutions of the Similarity estimation layer. The feature channels are compressed from $3C$ to $C$. Then we perform $\sum_{c=0}^{C}(\cdot)$ operation on the 4D volume to obtain a 3D volume. Finally we normalize the 3D volume by $softmax$ to produce the score volume $s(\cdot)$.


\paragraph{Scale uncertainty (SU)} 
SU represents the matching scale, which meets the requirements that coarse depth with large errors needs a large regression scale affording a large searching range to correct the error, and large depth requires dense matching for a more accurate estimation.  Therefore we learn SU from the coarse depth and the error of the coarse depth.

We subtract $f_{l,o}$ from the warped $\hat{f}_{l,o}$ (which is warped according to the coarse depth) to obtain the uncertainty feature $f_u$, which contains the reconstruction inconsistency error. Then the uncertainty feature is concatenated with coarse depth and sent to three 2D convolutions (Scale Uncertainty layer), obtaining the SU map. In this process, $f_u$ provides the error information and coarse depth provides the depth range information. At last, the SU is multiplied over the index $i$ to adjust its scale as in Equation \ref{offset}.

The SU and the aforementioned FU are interdependent. Different SU will lead to different matching granularity, and the FU will be adjusted correspondingly. 
\begin{table*}[h]
	\centering
	\caption{The ablative depth results of different components in DDL-Net on SceneFlow datasets. The baseline uses disparity-based cost volume and is supervised by disparity. BL: Baseline. Dep: use depth-based cost and supervised by depth. Unit: m.}
	\begin{tabular}{|c|c|c|c|c|c|c|c|c|l|}
		\hline
		Model&$[1,10)$ &$[10,20)$& $[20,30)$& $[30,40)$& $[40,50)$& $[50,60)$& $[60,70)$& $[70,80)$&MAE \\
		\hline
		Baseline &0.21&0.58&1.58&8.18&7.06&9.42&14.41&18.76&0.41 \\
		
		BL+Dep&0.21&0.36&0.81&3.30&4.66&5.54&8.13&9.92&0.29 \\
		
		BL+Dep+GU (DDL-Net)&\textbf{0.14}&\textbf{0.29}&\textbf{0.63}&\textbf{2.41}&\textbf{2.54}&\textbf{3.09}&\textbf{4.41}&\textbf{5.29}&\textbf{0.21}\\
		\hline
	\end{tabular}
    \label{ablation}
\end{table*}
\begin{table*}[h]
	\centering
	\caption{Depth accuracy comparison of different methods on Sceneflow dataset \cite{Sceneflow}. Unit: m.}
	\begin{tabular}{|c|c|c|c|c|c|c|c|c|l|}
		\hline
		Model& $[1,10)$ &$[10,20)$& $[20,30)$& $[30,40)$& $[40,50)$& $[50,60)$& $[60,70)$& $[70,80)$&MAE\\
		\hline
		PSM-Net (2018) \cite{PSMNet} &0.25&0.59&2.13&13.7&8.64&10.62&13.73&15.68&0.48\\
		GA-Net (2019) \cite{GANet}&0.15&0.48&1.66&7.71&5.58&7.43&9.75&11.25&0.33\\
		Gwc-Net (2019) \cite{GwcNet}&0.14&0.37&1.18&10.13&9.38&11.38&15.49&18.17&0.29\\
		AA-Net (2020) \cite{AANet}&0.18&0.52&1.84&7.27&5.75&7.68&10.15&11.85&0.38\\
		
		Bi3D-Net (2020) \cite{Bi3D}&0.47&1.24&8.05&11.76&3.59&5.77&7.50&9.69&0.77\\
		
		DeepPruner (2019) \cite{DeepPruner}&\textbf{0.11}&0.42&1.24&5.61&6.59&8.02&11.34&13.85&0.28\\
		\hline
		Ours&0.14&\textbf{0.29}&\textbf{0.63}&\textbf{2.41}&\textbf{2.54}&\textbf{3.09}&\textbf{4.41}&\textbf{5.29}&\textbf{0.21} \\		\hline
	\end{tabular}
    \label{sceneflow}
\end{table*}
\begin{table*}[h]
	\centering
	\caption{Depth accuracy comparison of different methods on DrivingStereo dataset \cite{drivingstereo}. Unit: m.}
	\begin{tabular}{|c|c|c|c|c|c|c|c|c|l|}
		\hline
		Model& $[1,10)$ &$[10,20)$& $[20,30)$& $[30,40)$& $[40,50)$& $[50,60)$& $[60,70)$& $[70,80)$&MAE \\
		\hline
		Gwc-Net (2019) \cite{GANet}&\textbf{0.08}&\textbf{0.18}&0.45&0.79&1.24&1.77&2.42&3.08&0.55 \\
		AA-Net (2020) \cite{AANet}&0.10&0.22&0.51&0.92&1.50&2.23&3.20&4.11&0.68 \\
		\hline
		UCS-Net (2020) \cite{UCSNet}&0.32&0.49&1.86&3.28&5.18&7.83&12.12&18.92&2.89 \\
		\hline
		Ours&0.13&0.20&\textbf{0.43}&\textbf{0.70}&\textbf{1.02}&\textbf{1.39}&\textbf{1.88}&\textbf{2.31}&\textbf{0.48} \\		
		\hline
	\end{tabular}
    \label{driving}
\end{table*}

\subsection{Loss function}
In order to directly estimate depth instead of disparity from the binocular camera system, we also directly utilize depth for supervision of both CDE and AGDR.  We denote the predicted depth as $depth$ and the depth ground-truth as $depth_{gt}$, $L_{1}$ loss can be represented as the following:
\begin{equation}
L_{1} = \frac{1}{N}\sum|depth-depth_{gt}|.
\end{equation}
\section{Experiments}
\begin{figure}[h]
\begin{center}
\includegraphics[width=0.47\textwidth]{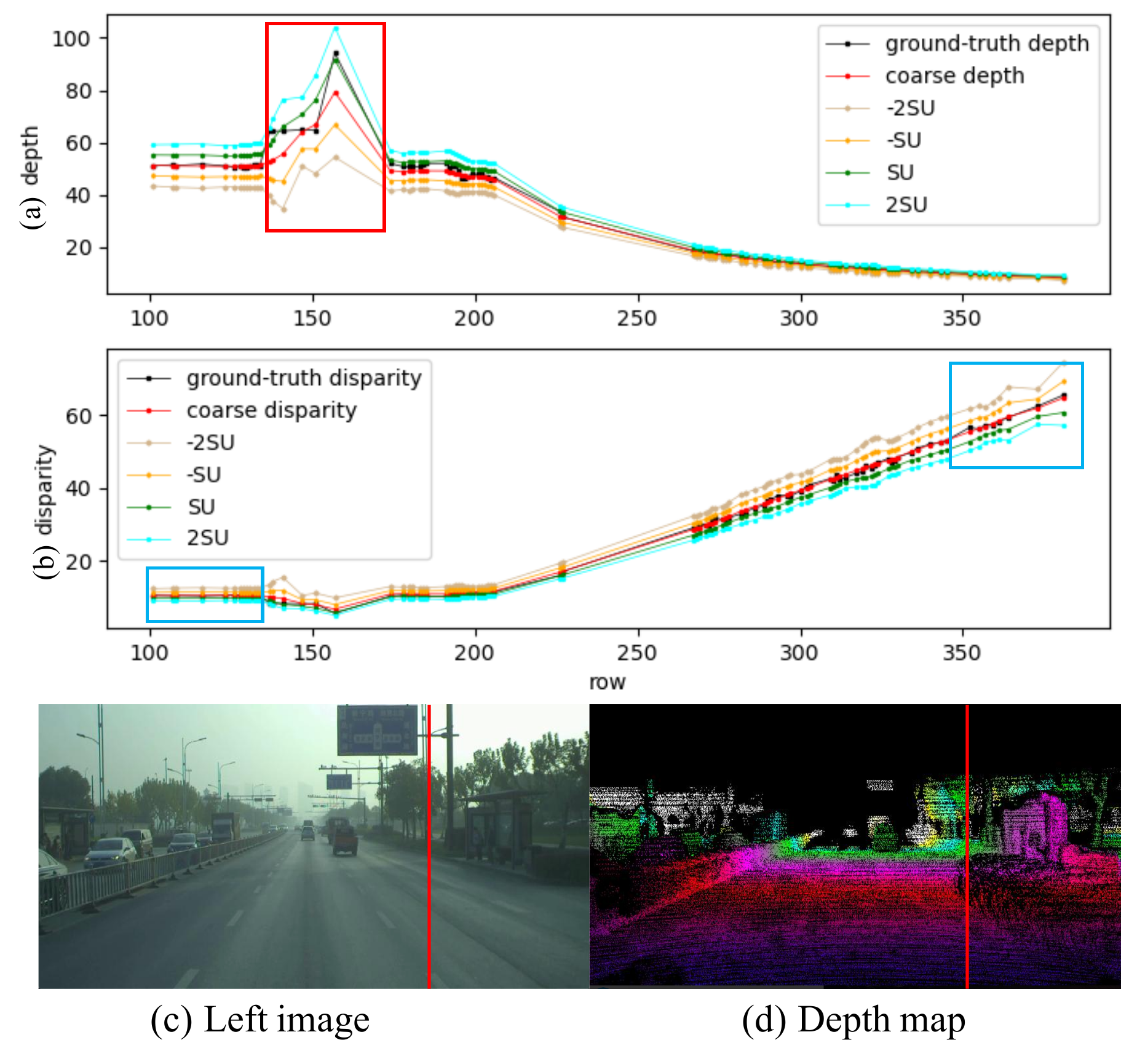}
\end{center}
   \caption{Visualization of the fine matching range decided by SU. The line graphs (a) (b) show the data of the selected column corresponding to red line in (c). -2SU, -SU, coarse depth, SU and 2SU are the positions of the candidates selected by SU. (a) draws the candidates' positions in the depth range, showing the matching range. (b) draws the candidates' positions in the right image, showing the matching granularity.}
\label{fig:scaleline}
\end{figure}
\begin{figure*}[h]
\begin{center}
\includegraphics[width=0.95\textwidth]{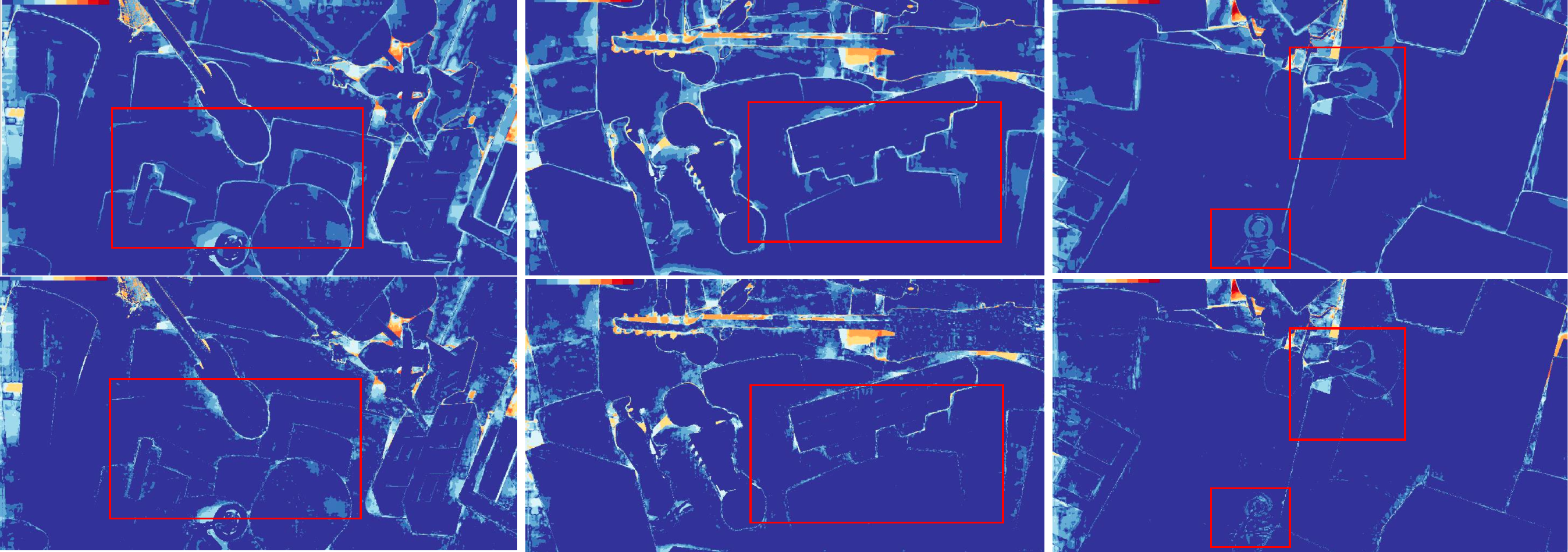}
\end{center}
   \caption{Visualization of the coarse depth (top) and the corresponding final depth (bottom) on SceneFlow dataset. Darker blue represents lower error.}
\label{fig:sceneflow}
\end{figure*}
\begin{figure*}[h]
\begin{center}
\includegraphics[width=0.95\textwidth]{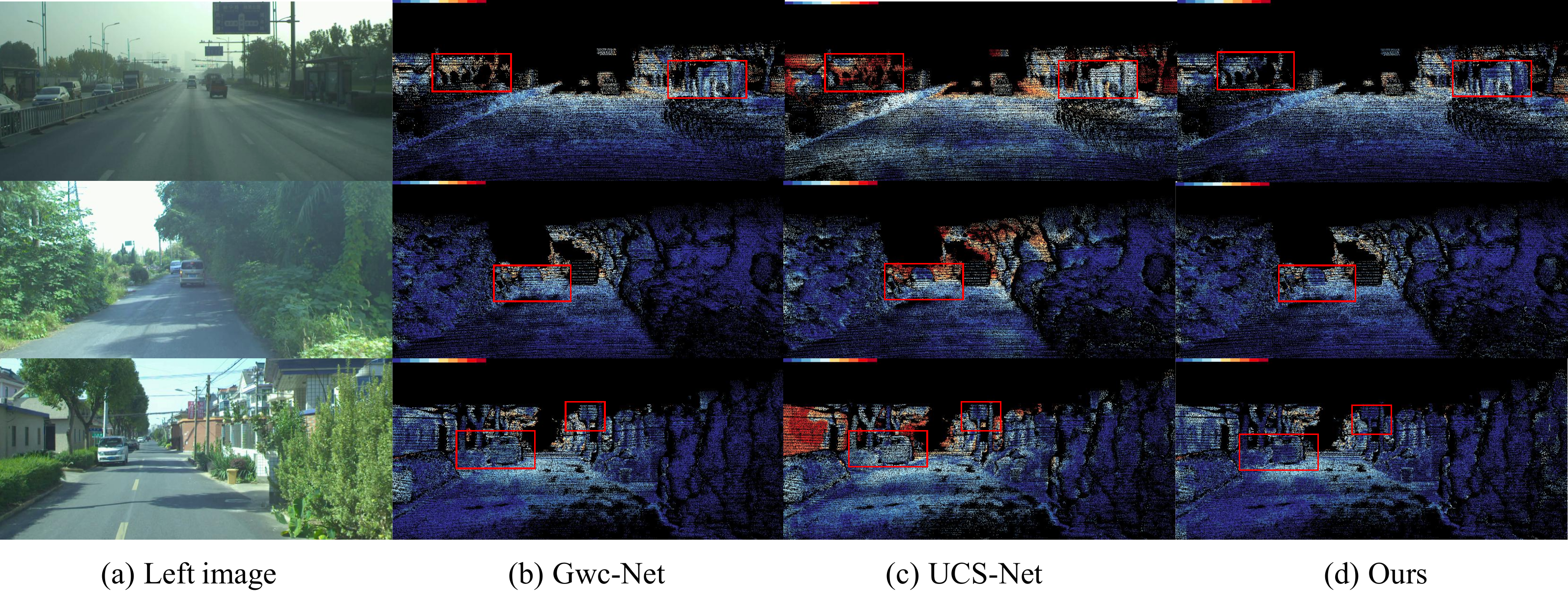}
\end{center}
   \caption{Visualization comparison of different methods on DrivingStereo dataset. Darker blue represents lower error.}
\label{fig:drivingstereo}
\end{figure*}

\subsection{Datasets and metrics}
\paragraph{Datasets.}We evaluate our methods on two main datasets. (1) Sceneflow dataset \cite{Sceneflow} is large scale synthetic dataset providing dense disparity ground-truth. The baseline and focal length are 27 cm and 1050 pixels, respectively. We convert the disparity into depth for training and testing. We train all models on the training set with 35454 stereo pairs and evaluate on the standard test set with 4370 stereo pairs.  (2) DrivingStereo \cite{drivingstereo} is a real-word dataset covering a diverse set of driving scenarios. It provides 180k image pairs with high-quality labels of disparity and depth. The baseline and focal length are 54 cm and 1003 pixels, repectively. We use the depth ground-truth for experiments. Because the dataset is too large, we randomly select 34888 image pairs for training and 6645 image pairs for validating. It is noted that we do not compare our methods on KITTI 2015 dataset \cite{kitti15}, a popular real-world dataset, because the evaluation server of the KITTI 2015 only provides disparity accuracy evaluation for the provided 200 testing image pairs and we have no access to these ground-truths.

\paragraph{Metrics.} We use mean absolute error (\emph{MAE}) to evaluate the integral level of depth estimation, and $MAE=\frac{1}{N}\sum|depth_{gt}-depth_{pred}|$. We use piecewise mean absolute error (\emph{PMAE}) to evaluate the depth accuracy in different depth range. \emph{PMAE} divides the depth range into several intervals and measure MAE within each interval.

\subsection{Implementation Details}
DDL-Net is implemented with Pytorch and on 2080Ti GPUs. We train the network on 4 GPUs with a batchsize of 8. We use Momentum SGD with the momentum as 0.8 and an initial learning rate of 0.001. The learning rate is adjusted by \emph{CosineAnnealing} with $T_{max}$ and $eta_{min}$ set as $5$ and $4e$-$8$ respectively. In the training phase, images in all datasets are cropped to $256\times 512$. The whole training process is performed in three phases. In the first phase, we only optimize the backbone with Scale Uncertainty layer and Feature Uncertainty layer fixed. The SU is set as $5m$ and the FU is set as $1$ for equal importance of each point. In the second phase, we train the backbone and the Scale Uncertainty layer jointly. In the last phase, we optimize the whole network. For Sceneflow, each phase is trained for 12, 8, and 24 epochs, respectively. For DrivingStereo, the 3 phases are trained for 20, 5, and 15 epochs, respectively.  
We also implement some classical methods as described in their papers exactly, and the analysis can be found in Section 4.4 and 4.5. For SceneFlow, we directly use the model weight trained by authors.  For DrivingStereo, we train and validate these methods on the selected training set and validating set, respectively.

\subsection{Ablation experiments}

We conduct ablation studies to understand the influence of different components in our proposed method. We design different runs on Sceneflow dataset. We select the two-stage method as the baseline (Baseline), which samples matching candidates according to the disparity range and utilizes disparity ground-truth as supervision. As shown in Table \ref{ablation}, the depth errors increase rapidly as the depth increases, because the disparity-based sampling method leads to an insufficient density of the matching candidates at a large distance where the correct matching point maybe missed.

We change the sampling method and uniformly sample matching candidates in the depth range instead of the disparity range (BL+Dep). The supervision information is changed to the depth at the same time. As shown in Table \ref{ablation}, when the depth is greater than 20m, the accuracy of distant points improves significantly, and the accuracy is improved more than $30\%$. The reason is that ``BL+Dep" selects denser candidates at farther distance than the ``Baseline", and fewer points are missed. 

However, the depth accuracy is still not ideal, because the refinement stage depends on the coarse depth, which may result in wrong or inappropriate matching range for the fine matching process. We introduce ``GU" into ``BL+Dep", as shown in Table \ref{ablation}, the accuracy is further improved more than $19\%$ both near and far, which demonstrates that GU successfully adjusts the matching granularity of the fine matching stage. 

\subsection{Accuracy comparison}

Our comparisons focus on depth accuracy. Table \ref{sceneflow} and Table \ref{driving} show the performance of different methods on SceneFlow and DrivingStereo dataset, respectively. We divide existing methods into two categories:

(1) Disparity estimation methods (PSM-Net, Gwc-Net, GA-Net, AA-Net, Bi3D-Net, DeepPruner), which focus on improving the performance of the cost aggregation network to improve the disparity accuracy. These methods all achieve high disparity accuracy and therefore, they have high accuracy at a small depth. For example, as shown in Table \ref{driving}, Gwc-Net has higher accuracy than DDL-Net at the depth range of $[1,20)$, and AA-Net has higher accuracy at the range of $[1,10)$. However, DDL-Net has higher depth accuracy at a large depth than all other methods in both the synthetic dataset and the real-world dataset, since we focus directly on the depth accuracy.

(2) Two-stage multi-view stereo methods (UCS-Net). UCS-Net estimates a new matching range according to the confidence of the coarse depth for the fine matching process. However, this method is designed for single model reconstruction, thus the depth range of which is more narrow than the driving scene. For more accurate depth accuracy at the large distance, the matching granularity is required to adapt to the depth range. Therefore, we estimate GU to adjust the matching granularity to be denser for a large depth and the matching range to be wider for large errors. Finally, we have better performance at a different distance. As shown in Table \ref{driving}, we have more than $50\%$ gain than UCS-Net.

It is noted that the baselines and depth distributions for DrivingStereo dataset and SceneFlow dataset are different, thus the estimated depth errors for two datasets differ evidently.

\subsection{Visualization analysis}
 Firstly, we visualize the learned SU on DrivingStereo dataset. In this experiment, we intend to demonstrate that the learned SU satisfies the requirements: providing a larger matching range for larger error and denser matching granularity for larger depth. As shown in Figure \ref{fig:scaleline}, in the red box of (a), a large SU is learned for large depth errors. Accordingly, for small errors out of the red box, the matching range varies with the depth range. The granularity of the selected matching candidates on the right image can be seen from (b). In the blue boxes of (b), when the distance between the red line and the black line is similar (which means similar errors), we can see denser matching granularity is selected for far points while sparse granularity for near points.
 
 Secondly, we visualize the error maps of the depth before and after AGDR on SceneFlow datasets. As shown in the coarse depth error map of Figure \ref{fig:sceneflow} (top row), large errors concentrate on the occlusion and edge areas, and the error map appears in light blue in these areas. After AGDR, the light blue areas are reduced (bottom row). This demonstrates the effectiveness of the proposed AGDR. 
 
 Finally, we compare the error map with other methods on DrivingStereo dataset. As shown in the red boxes of Figure \ref{fig:drivingstereo}, DDL-Net has lower errors in the large depth. The results demonstrate that focusing on depth accuracy and adjusting the matching granularity according to the confidence and the depth range simultaneously do maintain high accuracy at near and achieve higher accuracy at far.

\section{Discussion and conclusion}
Stereo matching has been extensively exploited in the context of autonomous driving. Both traditional methods and learning-based stereo matching methods estimate disparity first. The farther the point is, the more sensitive its depth accuracy is to sub-pixel level disparity error. In practice, in order to obtain more accurate depth, the baseline of stereo cameras will be increased to reduce the influence of sub-pixel error.  However, for the requirement of the overlap area in stereo images, the baseline cannot be infinitely enlarged. Therefore, it is necessary to directly estimate depth to reduce the influence of baseline.
On the other hand, existing research focuses too much on ranking on Benchmarks. They put more effort into improving the accuracy of disparity while ignore the accuracy of depth. However, it is the depth that will be employed in practice. Therefore, more attention should be paid to improve depth accuracy directly.

In this paper, we analyze the above problems and propose a new framework for direct depth estimation. The proposed framework is supervised by depth and output depth. We also propose GU to adjust the fine matching stage to adapt to the depth value and the confidence of the coarse prediction simultaneously. The SU of GU adjusts matching range by changing the scale of the offset. The FU of GU adjusts the matching granularity by selecting the necessary matching candidates. We not only maintain the accuracy of near points but also improve the depth accuracy at far significantly.


{\small
\bibliographystyle{ieee_fullname}
\bibliography{egbib}
}

\end{document}